\documentclass{article}
\usepackage{spconf,amsmath,graphicx,subfigure,amssymb}

\title{Associating Multi-Scale Receptive Fields for Fine-grained Recognition}
\name{Zihan Ye$^{1}$, Fuyuan Hu$^{1,2*}$\thanks{* Corresponding author: fuyuanhu@mail.usts.edu.cn. This work was supported by NSFC (Nos. 61876121, 61728205, 61502329 and 61672371), Primary Research \& Development Plan of Jiangsu Province (No. BE2017663), and Jiangsu Key Disciplines off Thirteen Five-Year Plan (No. 20168765).}, Yin Liu$^{4}$, Zhenping Xia$^{1,3}$, Fan Lyu$^{5}$, Pengqing Liu$^{1}$}
\address{$^{1}$Suzhou University of Science and Technology\\
	$^{2}$Virtual Reality Key Laboratory of Intelligent Interaction and Application Technology of Suzhou\\
	$^{3}$Suzhou Key Laboratory for Big Data and Information Service\\
	$^{4}$Shanghai Institute of Technology\\
	$^{5}$Tianjin University
}
\begin{document}
%
\maketitle
\begin{abstract}
	Extracting and fusing part features have become the key of fined-grained image recognition.
	Recently, Non-local (NL) module has shown excellent improvement in image recognition.
	However, it lacks the mechanism to model the interactions between multi-scale part features, which is vital for fine-grained recognition.
	In this paper, we propose a novel cross-layer non-local (CNL) module to associate multi-scale receptive fields by two operations.
	First, CNL computes correlations between features of a query layer and all response layers.
	Second, all response features are weighted according to the correlations and are added to the query features.
	Due to the interactions of cross-layer features, our model builds  spatial dependencies among multi-level layers and learns more discriminative features.
	In addition, we can reduce the aggregation cost if we set low-dimensional deep layer as query layer.
	Experiments are conducted to show our model achieves or surpasses state-of-the-art results on three benchmark datasets of fine-grained classification.
	Our codes can be found at \emph{github.com/FouriYe/CNL-ICIP2020}.
\end{abstract}
\section{Introduction}
\label{sec:intro}

\begin{figure}[htb]
	\begin{minipage}[b]{1.0\linewidth}
		\centering
		\centerline{\includegraphics[width=.8\linewidth]{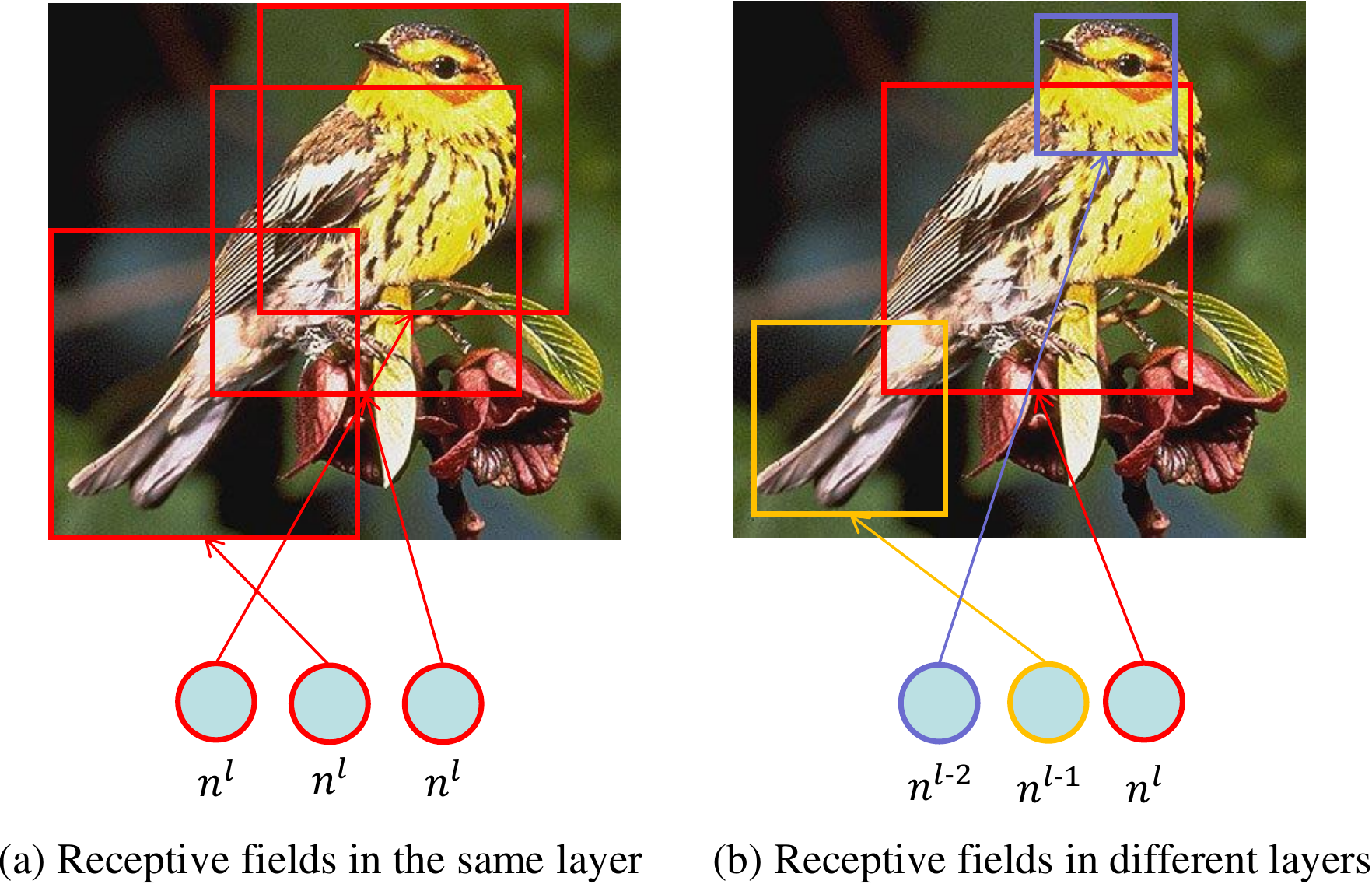}}
		\vspace{-0.2cm}
	\end{minipage}
	\caption{Receptive fields of corresponding neurons.
		We denote the neurons in $l$-th layer as $n^{l}$.
		Obviously, when a receptive field is over-sized, it contain more background region than a appropriately sized one.}
	\label{fig:RF1}
	\vspace{-0.5cm}
\end{figure}

Fine-grained image classification focuses on categorizing very similar subgroups, e.g. the Husky and Alaska in dogs.  
It has long been considered as a challenging task due to discriminative region localization and discriminative feature learning from those regions.
Early works on localization-based methods often adopt part detectors trained on part annotations and then extract part-specific features for fine-grained classification \cite{xu2015augmenting, xiao2015application,simon2015neural,zhang2016picking}.
However, these strongly-supervised methods heavily rely on manual object part annotations, which is too expensive to be prevalently applied in practice.
Therefore, the attention mechanism~\cite{fu2017look,sun2018multi, ye2019unsupervised,ye2018daugan} have received increasing attention in recent researches, which can be implemented as one kind of weakly-supervised frameworks that do not need part annotations.
\cite{fu2017look} proposes a recurrent attention convolutional neural network that progressively localizes discriminative areas and extracts features from coarse to fine scale.
\cite{sun2018multi} proposes a soft attenion-based model that generates semantic region features.

Recently, non-local (NL) module~\cite{wang2018non} is proposed to model the long-range dependencies in one layer, via self-attention mechanism to form an attention map.
For each position, the NL module first computes the pairwise relations between current position and all positions, and then aggregates the features of all positions by weighted sum.
The aggregated features are added to the feature of each position to form the output.
\cite{yue2018compact,cao2019gcnet} show the NL module could bring excellent improvement in image recognition.
However, for fine-grained classification task, multi-scale features are vital because object parts have various sizes and shapes in images~\cite{zhao2017diversified,luo2019cross,he2015spatial}.
One NL module models the spatial dependencies in only one convolutional layer, in which neurons have fixed-size receptive fields.
The mismatch between the sizes of receptive fields and parts of objects might undermines the feature extract.
For example, as shown in Fig.\ref{fig:RF1} (a), an over-sized receptive fields contain more background region, and sequentially bring more noises.
The noises brought by meaningless background have been proven to be harmful for classification~\cite{zhang2018occluded,luo2015multiview}.

To address the issue, we propose a novel Cross-layer Non-Local module (CNL).
We first decompose traditional non-local module into two basic components, query layer and response layer.
Then, based on the observation that neurons in different layers has various receptive fields as shown in Fig.\ref{fig:RF1} (b), we associate one query layer with several response layers.
Consequently, the query layer could learn more discriminative multi-scale feature from distant response layers.
The explicitly learned correlations among all related layers improve the representation power of network.

To the best of our knowledge, we are the first that explore the non-local operation into cross-layer feature fusion for fined-grained classification.
We certify that the cross-layer operation could ease the computation complexity of associating receptive fields in theory.
We evaluate our CNL method on three datasets of fine-grained classification, i.e. CUB-200-2011~\cite{wah2011caltech}, Stanford Dogs~\cite{khosla2011novel} and Stanford Cars~\cite{he2016deep}.
Experimental results show that: (1) as a plug-and-play module, CNL needs less cost than NL module and other methods, but surpasses or achieves state-of-the-art performance,  (2) CNL can adaptively focus on various parts in multi-scale.

\section{Approach}
\label{sec:approach}

In this section, we first review the original non-local (NL) module.
Then, we introduce a general formulation of the NL module and show that the NL is a special case of the formulation.
After that, we propose our Cross-layer Non-Local module (CNL) for multi-scale receptive fields association.

\subsection{Review of Non-local Module}

\begin{figure}[htbp]
	\centering
	
	\begin{minipage}[t]{\linewidth}
		\centering
		\includegraphics[width=\linewidth]{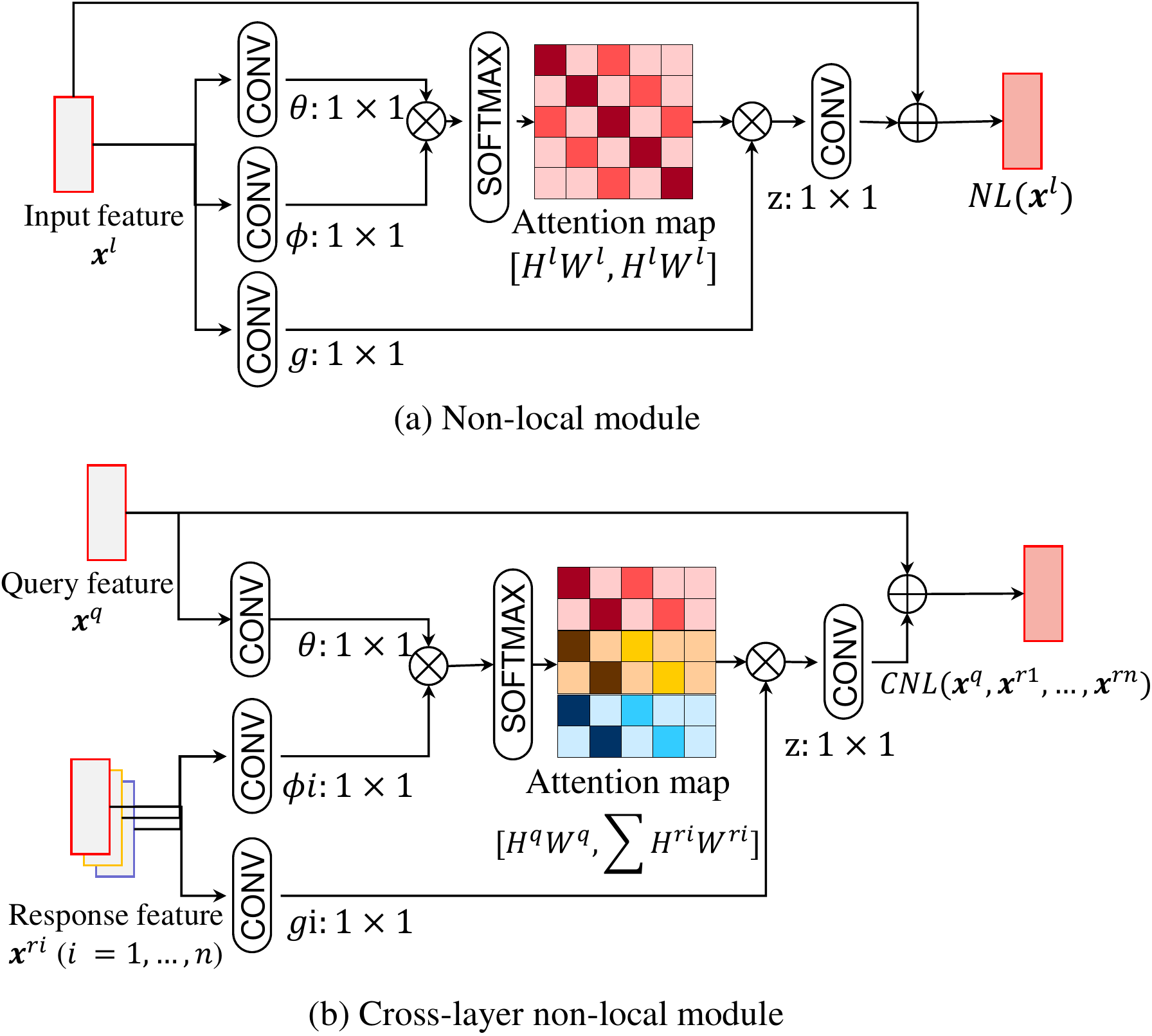}\\
	\end{minipage}%
	\centering
	\vspace{-0.2cm}
	\caption{Architecture of NL module and proposed CNL module. $\bigotimes$ is matrix multiplication, and $\bigoplus$ is element-wise addition. The attention maps are shown by their dimensions, e.g. $[HW,HW]$. NL module associates single-scale features in only one layer. Our proposed CNL module associates the query layer with multi-level response layers to model the dependencies of multi-scale features.}
	\label{fig:methods}
	\vspace{-0.5cm}
\end{figure}

For a feature map in the $l$-th convolutional layer, $H^{l}$, $W^{l}$ and $C^{l}$ denotes its height, width and the number of channels, respectively.
And we denote the feature map as $\mathbf{X}^{l}\in \mathbf{R}^{H^{l}W^{l}\times C^{l}}$(see notation\footnote{Bold capital letters denote matrices, all non-bold letters represent scalars. Superscripts of features denote indices of corresponding layers.}).
The row vector of the feature map could be considered as a feature vector extracted from corresponding receptive field.
To capture long-range dependencies across the whole feature map $\mathbf{X}^{l}$, the original non-local operation first uses two learnable transformations, $\theta(\cdot)$ and $\phi(\cdot)$, to project $\mathbf{X}^{l}$ into a embedding space.
And next, the attention map is computed by a pairwise function $f(\cdot,\cdot)$ in the embedding space.
The value of every element in the attention map indicates   the affinity of corresponding feature vectors.
Finally, the features at all positions are projected by another learnable transformations $g(\cdot)$.
In addition, to reduce computation costs, $\theta(\cdot)$, $\phi(\cdot)$ and $g(\cdot)$ would shrink the channel of input feature.
The convolution $z$ is to maintain the channel consistency between input feature $\mathbf{X}^{l}$ and processed feature $NL(\mathbf{X}^{l})$.
The final result is the weighted sum of the embedding features at all positions: 
\begin{equation}
	NL(\mathbf{X}^{l}) = \mathbf{X}^{l} + z(f(\theta(\mathbf{X}^{l}),\phi(\mathbf{X}^{l}))g(\mathbf{X}^{l})). 
\end{equation}
There are multiple choices for $f$ suggested in\cite{wang2018non}.
The dot-product is generally chosen to implement $f$.
$\theta(\cdot)$, $\phi(\cdot)$ and $g(\cdot)$ are generally implemented as $1\times 1$ convolution.

\subsection{Cross-layer Non-local Module}
The NL module is one kind of the self-attention module\cite{vaswani2017attention}.
We can indicate the input feature $\mathbf{X}^{l}$ as two different symbols in two data flows, respectively, according to their functions.
One flow is only inputted to $\theta$ convolution in order to compute the attention map.
We denote the feature in this flow as $\mathbf{X}^{q}\in \mathbf{R}^{H^{q}W^{q}\times C^{q}}$.
The other flow is not only inputted to $\phi(\cdot)$ for the attention map, but also is inputted to $g(\cdot)$ for the response results.
We denote this feature as $\mathbf{X}^{r}\in \mathbf{R}^{H^{r}W^{r}\times C^{r}}$.
In fact, we can rewrite it as a more general two-variables form: 
\begin{equation}
NL_{bi}(\mathbf{X}^{q},\mathbf{X}^{r}) = \mathbf{X}^{l} +  z(f(\theta(\mathbf{X}^{q}),\phi(\mathbf{X}^{r}))g(\mathbf{X}^{r})).
\end{equation}
Obviously, the formulation of original NL module is a special case of the two-variables form.
In that case, the query layer $q$ is equal to the response layer $r$.

Deep layers are sensitive for large parts since their large-scale receptive fields.
Shadow layers do the opposite.
According to the fact, we directly compute the affinity among deep features and shadow features.
Specially, we consider one deep layer as query layer and associate it with several shadow response layers.
In this association way, our model is enhanced in mainly two ways.
First, our decision layer could learn more discriminative multi-scale features that may be impacted by meaningless background noises in deep layers.
Second, our model could explicitly compute the dependencies of multi-scale image regions.

Now, we give formal definition of our \textbf{C}ross-layer \textbf{N}on-\textbf{L}ocal module (CNL).
Given $n$ response layers, we denote $r1, r2,\cdots, rn$ as the indices of those layers from shallow to deep.
Our CNL operation is expressed as:
\begin{equation}
	CNL(\mathbf{X}^{q}, \mathbf{X}^{r1}, \cdots, \mathbf{X}^{rn}) = \mathbf{X}^{q} + \sum^{}_{} \mathbf{Z}^{ri},
	\vspace{-0.4cm}
\end{equation}
where $\mathbf{Z}^{ri} = z(f(\theta(\mathbf{X}^{q}),\phi i(\mathbf{X}^{ri}))gi(\mathbf{X}^{ri}))$.

The NL module has a well-known quadratic space complexity, which is mainly from the stored attention map, i.e. $\mathcal{O}((H^{l}W^{l})^{2})$.
Obviously, the NL module cost is immense when insert it into a shadow layer for low-level vision tasks.
For one extreme example, the resolution of the first layer in ResNet-50~\cite{he2016deep,wang2018non} is $112\times112$ for a $224\times224$ image. when we insert NL module after that layer, we need store a $112^2\times 112^2$ attention map, which is unaffordable.
Our CNL module avoid the problem.
We directly associate shadow feature in deep feature space.
In other words, we set the query layer in the last convolutional layer of ResNet-50, in which feature map has a $7 \times 7$ resolution.
In that case, the size of our attention map  has shrunk to $(112*7)\times (112*7)$.
It is a discount of about \textbf{99.4\%} .
Section\ref{sec:ablation} shows controlled experiments between CNL and NL module in practical cases.

\section{experiment}
\label{sec:experiment}

\begin{table}
	\caption{The statistics of fine-grained datasets used in this paper.}
	\label{table:datasets}
	\centering
	\begin{tabular}{p{2.9cm}|p{1.4cm}|p{1.5cm}|p{1.3cm}}
		\hline
		Datasets & \#Category & \# Training & \# Testing\\
		\hline
		CUB-200-2011~\cite{wah2011caltech} & 200 & 5,994 & 5,794\\
		Stanford Dogs~\cite{khosla2011novel} & 120 & 12,000 & 8,580\\
		Stanford Cars~\cite{he2016deep} & 196 & 8,144 & 8,041\\
		\hline
	\end{tabular}
\vspace{-0.5cm}
\end{table}

\begin{table}
	\caption{Experimental results of ablation. Top1 and top5 accuracy (\%) on CUB-200-2011.}
	\label{table:ablation}
	\centering
	\begin{tabular}{p{1.8cm}|p{0.8cm}|p{0.8cm}|p{1.82cm}|p{1.7cm}}
		\hline
		Method & top1 & top5 & \#param($\times 10^{7}$) & Flops($\times 10^{9}$) \\
		\hline
		ResNet50 & 84.05  & 96.00 & 2.39 & 1.64\\
		+5NL     & 85.10 & 96.18 & 3.13 & 2.48\\
		+5CNL(ours)  & 85.64 & 96.84 & 2.71 & 2.03 \\
		\hline
		ResNet101  & 85.05 & 96.70 & 4.29 & 3.15 \\
		+5NL  & 85.53 & 96.65 & 5.03 & 3.96 \\
		+5CNL(ours)  & 86.73 & 96.75 & 4.61 & 3.52 \\
		\hline
	\end{tabular}
\vspace{-0.25cm}
\end{table}

\begin{figure}[htb]
	\begin{minipage}[b]{1.0\linewidth}
		\centering
		\centerline{\includegraphics[width=\linewidth]{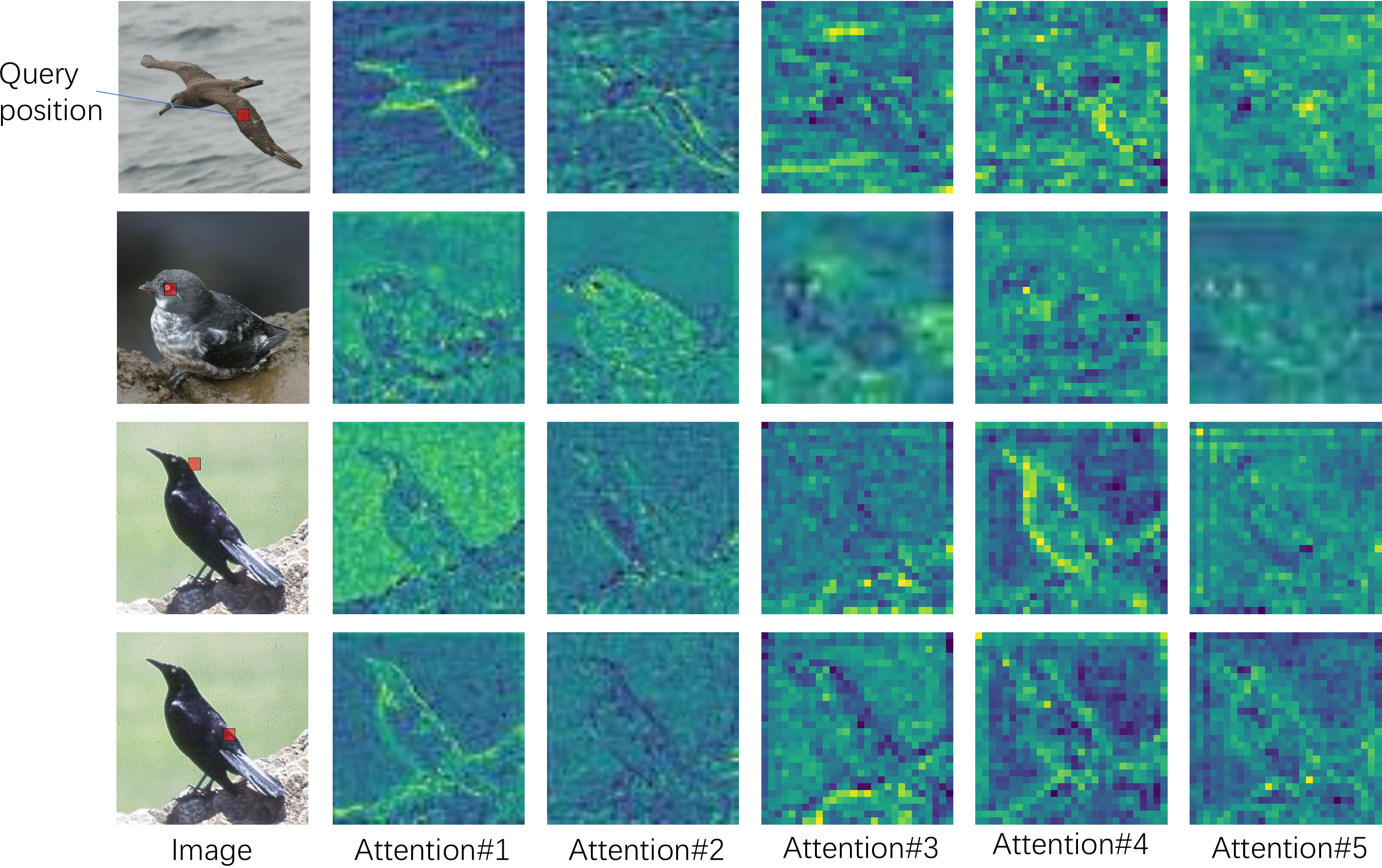}}
	\end{minipage}
	\caption{Visualization with attention maps for different query positions (red blocks) in CUB fine-grained classification. Column 1: the input image. Column 2 to 6: attention maps from shadow response layers to deep response layers.}
	\label{fig:vis}
	\vspace{-0.5cm}
\end{figure}

\subsection{Datasets and training details}
We conduct experiments on three challenging fine-grained image classification datasets, i.e. CUB-200-2011~\cite{wah2011caltech}, Stanford Dogs~\cite{khosla2011novel} and Stanford Cars~\cite{he2016deep}. The detailed statistics with category numbers and data splits are summarized in Table~\ref{table:datasets}.
In our experiments, the input images are resized to $448\times448$ for both training and testing.
We train each dataset for 110 epochs with the gradual warmup strategy~\cite{he2016deep} used in first ten epochs.
The batch size is set to 64 and the base learning rate is set to 0.01, with decays by 0.9 for the $50th$, $70th$ and $90th$ epoch.
Our method is implemented with PyTorch and one RTX 8000 GPU.
We adopt the ResNet~\cite{he2016deep} series (ResNet50 and ResNet101) as our candidate baselines.
We use the models pretrained on ImageNet~\cite{deng2009imagenet} to initialize the weights.
To keep the same setting with~\cite{wang2018non}, we use zero to initialize the weight and bias of the BatchNorm (BN) layer in both NL and CNL blocks~\cite{goyal2017accurate}.

\begin{table*}
	\caption{Comparison results on CUB-200-2011, Stanford Dogs and Stanford Cars datasets. "Anno." stands for using extra annotation(bounding box or part) in training. "1-Stage" indicates whether the training can be done in one stage.}
	\centering
	\label{table:state}
	\begin{minipage}{0.5\linewidth}
		\label{table:cub}
		\resizebox{0.9\linewidth}{!}{
		\begin{tabular}{p{3.7cm}|p{0.8cm}|p{1.2cm}|p{0.8cm}}
			\multicolumn{4}{c}{(a) CUB-200-2011}\\
			\hline
			Method & Anno. & 1-Stage & top1 \\
			\hline
			Part-RCNN \cite{zhang2014part} &$\checkmark$& $\times$ & 81.6\\
			PN-CNN \cite{branson2014bird} &$\checkmark$& $\times$ & 85.4\\
			\hline
			DVAN \cite{zhao2017diversified} &$\times$& $\times$ & 79.0\\
			B-CNN \cite{lin2015bilinear} &$\times$& $\times$ & 84.1\\
			PDFR \cite{zhang2016picking} &$\times$& $\times$ & 84.5\\
			RACNN \cite{fu2017look} &$\times$& $\times$ & 85.3\\
			RAM \cite{li2017dynamic} &$\times$& $\times$ & 86.0\\
			\hline
			DeepLAC \cite{lin2015deep} &$\checkmark$& $\checkmark$ & 80.3\\
			PA-CNN \cite{krause2015fine} &$\checkmark$& $\checkmark$ & 82.8\\
			SPDA-CNN \cite{zhang2016spda} &$\checkmark$& $\checkmark$ & 85.1\\
			\hline
			NAC \cite{simon2015neural} &$\times$& $\checkmark$ & 81.0\\
			FCAN \cite{liu2016fully} &$\times$& $\checkmark$ & 84.7\\
			MACNN \cite{zheng2017learning} &$\times$&$\checkmark$&86.5\\
			MAMC \cite{sun2018multi} &$\times$& $\checkmark$ & 86.5\\
			Ours (ResNet101+5CNL) &$\times$& $\checkmark$ & \textbf{86.7}\\
			\hline
		\end{tabular}}
	\end{minipage}
	\begin{minipage}{0.48\linewidth}
		\centering
		\begin{minipage}{\linewidth}
			\resizebox{0.9\linewidth}{!}{
			\begin{tabular}{p{3.7cm}|p{0.8cm}|p{1.2cm}|p{0.8cm}}
				\multicolumn{4}{c}{(b) Stanford Dogs}\\
				\hline
				Method & Anno. & 1-Stage & top1 \\
				\hline
				PDFR \cite{zhang2016picking} &$\times$& $\times$ & 72.0\\
				DVAN \cite{zhao2017diversified} &$\times$& $\times$ & 81.5\\
				RACNN \cite{fu2017look} &$\times$& $\times$ & \textbf{87.3}\\
				\hline
				NAC \cite{simon2015neural} &$\times$& $\checkmark$ & 68.6\\
				FCAN \cite{liu2016fully} &$\times$& $\checkmark$ & 84.2\\
				MAMC \cite{sun2018multi} &$\times$& $\checkmark$ & 85.2\\
				Ours (ResNet101+5CNL) &$\times$& $\checkmark$ & 85.2\\
				\hline
			\end{tabular}}
		\end{minipage}
		
		\begin{minipage}{\linewidth}
			\vspace{0.1cm}
			\label{table:cars}
			\resizebox{0.9\linewidth}{!}{
			\begin{tabular}{p{3.7cm}|p{0.8cm}|p{1.2cm}|p{0.8cm}}
				\multicolumn{4}{c}{(c) Stanford Cars}\\
				\hline
				Method & Anno. & 1-Stage & top1 \\
				\hline
				DVAN \cite{zhao2017diversified} &$\times$& $\times$ & 87.1\\
				B-CNN \cite{lin2015bilinear} &$\times$& $\times$ & 91.3\\
				RACNN \cite{fu2017look} &$\times$& $\times$ & 92.5\\
				\hline
				PA-CNN \cite{krause2015fine} &$\checkmark$& $\checkmark$ & 92.8\\
				\hline
				FCAN \cite{liu2016fully} &$\times$& $\checkmark$ & 89.1\\
				MAMC \cite{sun2018multi} &$\times$& $\checkmark$ & 93.0\\
				Ours (ResNet101+5CNL) &$\times$& $\checkmark$ &\textbf{93.1}\\
				\hline
			\end{tabular}}
		\end{minipage}
	\end{minipage}
\vspace{-0.5cm}
\end{table*}

To investigate the receptive field association ability, we insert 5 modules in different layers .
Specially, we insert 2 NL modules in $res3$, 3 NL modules in $res4$ at regular interval.
For CNL module, we set the same layers as response layers, but set the last layer in $res5$ as query layer.

\subsection{Ablations}
\label{sec:ablation}

We first take the ablation experiments in CUB-200-2011.
Table~\ref{table:ablation} shows quantitative results of networks in different configurations.
For ResNet-50, using NL module can offer 1.05\% top1 accuracy imporvement compared to the baseline (85.10\% vs. 84.05\%).
With CNL module, the model boosts the accuracy by 0.54\% compared to that model with NL module.
The similar results could see in the other ablations for ResNet-100.
Benefiting from the association in the deep feature space, CNL module has less computation cost then NL module.
The ResNet-100 with CNL saves 0.74*10$^7$ parameters compared to that one using NL module.

\subsection{Comparison with State-of-the-Art}

We compare several state-of-the-art results to our methods in Table~\ref{table:state} on these three datasets.
For a intuitive comparison, we group those methods according to whether they use extra annotation and whether they use multiple training stages.
Our CNL can be trained end-to-end in one stage, and does not need any expensive part annotation.

In general, our method achieve the best score within its all groups.
In CUB and Stanford Cars, our method beat the second best method MAMC~\cite{sun2018multi} by 0.2\% and 0.1\% top1 accuracy, respectively.
It is worth to noted that our method only need 11.7\% additional Flops, but MAMC needs 23.4\% additional computation cost according to its paper.
In Stanford Dogs, our method achieve the second best performance.
Although RACNN has the excellent result in the dataset, it requires multiple training stages and is tough to be trained end-to-end.
Our CNL is easy to implemented, only need little cost, and could bring significant improvement.

%

%

\subsection{Visualization with Attenion Map}
To understand the effects of CNL, we visualize its attention maps as shown in Fig~\ref{fig:vis}. 
The brighter the pixel is, the more CNL cares about that area.
We can see two points: (1) our network can focus on meaningful semantic parts, e.g. sky, edge, eyes and body of birds, (2) our network can adaptively focus on different parts in different response layers.
For example, in the second row, the query position is in the eye of the bird. In the second attention map, the position of eye is less noticed than the body positions.
However, in the fourth attention map, the eye position receives the highest attention, which further validates the effectiveness of our method.
\vspace{-0.5cm}
\section{Conclusion}
\label{sec:conclusion}
In this paper, we have introduced a general formulation of Non-Local (NL) module, and have proposed a novel Cross-layer Non-Local (CNL) module for fined-grained image recognition.
Our model could learn more discriminative features and adaptively focuses on multi-scale image parts by associating cross-layer features.
Additionally, our CNL module eases the heavy association cost of original NL operation.
Our model needs less computational cost compared to other modules, but produces more competitive results and surpasses state-of-the-art results.

%

\bibliographystyle{IEEEbib}
\bibliography{strings,refs}

\begin{thebibliography}{10}

\bibitem{xu2015augmenting}
Zhe Xu, Shaoli Huang, Ya~Zhang, and Dacheng Tao,
\newblock ``Augmenting strong supervision using web data for fine-grained
  categorization,''
\newblock in {\em CVPR}, 2015.

\bibitem{xiao2015application}
Tianjun Xiao, Yichong Xu, Kuiyuan Yang, Jiaxing Zhang, Yuxin Peng, and Zheng
  Zhang,
\newblock ``The application of two-level attention models in deep convolutional
  neural network for fine-grained image classification,''
\newblock in {\em CVPR}, 2015.

\bibitem{simon2015neural}
Marcel Simon and Erik Rodner,
\newblock ``Neural activation constellations: Unsupervised part model discovery
  with convolutional networks,''
\newblock in {\em CVPR}, 2015.

\bibitem{zhang2016picking}
Xiaopeng Zhang, Hongkai Xiong, Wengang Zhou, Weiyao Lin, and Qi~Tian,
\newblock ``Picking deep filter responses for fine-grained image recognition,''
\newblock in {\em CVPR}, 2016.

\bibitem{fu2017look}
Jianlong Fu, Heliang Zheng, and Tao Mei,
\newblock ``Look closer to see better: Recurrent attention convolutional neural
  network for fine-grained image recognition,''
\newblock in {\em CVPR}, 2017.

\bibitem{sun2018multi}
Ming Sun, Yuchen Yuan, Feng Zhou, and Errui Ding,
\newblock ``Multi-attention multi-class constraint for fine-grained image
  recognition,''
\newblock in {\em ECCV}, 2018.

\bibitem{ye2019unsupervised}
Zihan Ye, Fan Lyu, Linyan Li, Yu~Sun, Qiming Fu, and Fuyuan Hu,
\newblock ``Unsupervised object transfiguration with attention,''
\newblock {\em Cognitive Computation}, 2019.

\bibitem{ye2018daugan}
Zihan Ye, Fan Lyu, Jinchang Ren, Yu~Sun, Qiming Fu, and Fuyuan Hu,
\newblock ``Dau-gan: Unsupervised object transfiguration via deep attention
  unit,''
\newblock pp. 120--129, 2018.

\bibitem{wang2018non}
Xiaolong Wang, Ross Girshick, Abhinav Gupta, and Kaiming He,
\newblock ``Non-local neural networks,''
\newblock in {\em CVPR}, 2018.

\bibitem{yue2018compact}
Kaiyu Yue, Ming Sun, Yuchen Yuan, Feng Zhou, Errui Ding, and Fuxin Xu,
\newblock ``Compact generalized non-local network,''
\newblock in {\em NeurIPS}, 2018.

\bibitem{cao2019gcnet}
Yue Cao, Jiarui Xu, Stephen Lin, Fangyun Wei, and Han Hu,
\newblock ``Gcnet: Non-local networks meet squeeze-excitation networks and
  beyond,''
\newblock in {\em CVPR Workshops}, 2019.

\bibitem{zhao2017diversified}
Bo~Zhao, Xiao Wu, Jiashi Feng, Qiang Peng, and Shuicheng Yan,
\newblock ``Diversified visual attention networks for fine-grained object
  classification,''
\newblock {\em TMM}, 2017.

\bibitem{luo2019cross}
Wei Luo, Xitong Yang, Xianjie Mo, Yuheng Lu, Larry~S Davis, Jun Li, Jian Yang,
  and Ser-Nam Lim,
\newblock ``Cross-x learning for fine-grained visual categorization,''
\newblock in {\em ICCV}, 2019.

\bibitem{he2015spatial}
Kaiming He, Xiangyu Zhang, Shaoqing Ren, and Jian Sun,
\newblock ``Spatial pyramid pooling in deep convolutional networks for visual
  recognition,''
\newblock {\em TPAMI}, 2015.

\bibitem{zhang2018occluded}
Shanshan Zhang, Jian Yang, and Bernt Schiele,
\newblock ``Occluded pedestrian detection through guided attention in cnns,''
\newblock in {\em CVPR}, 2018.

\bibitem{luo2015multiview}
Yong Luo, Tongliang Liu, Dacheng Tao, and Chao Xu,
\newblock ``Multiview matrix completion for multilabel image classification,''
\newblock {\em TIP}, 2015.

\bibitem{wah2011caltech}
Catherine Wah, Steve Branson, Peter Welinder, Pietro Perona, and Serge
  Belongie,
\newblock ``The caltech-ucsd birds-200-2011 dataset,''
\newblock 2011.

\bibitem{khosla2011novel}
Aditya Khosla, Nityananda Jayadevaprakash, Bangpeng Yao, and Fei-Fei Li,
\newblock ``Novel dataset for fine-grained image categorization: Stanford
  dogs,''
\newblock in {\em CVPR Workshops}, 2011.

\bibitem{he2016deep}
Kaiming He, Xiangyu Zhang, Shaoqing Ren, and Jian Sun,
\newblock ``Deep residual learning for image recognition,''
\newblock in {\em CVPR}, 2016.

\bibitem{vaswani2017attention}
Ashish Vaswani, Noam Shazeer, Niki Parmar, Jakob Uszkoreit, Llion Jones,
  Aidan~N Gomez, {\L}ukasz Kaiser, and Illia Polosukhin,
\newblock ``Attention is all you need,''
\newblock in {\em NeurIPS}, 2017.

\bibitem{deng2009imagenet}
Jia Deng, Wei Dong, Richard Socher, Li-Jia Li, Kai Li, and Li~Fei-Fei,
\newblock ``Imagenet: A large-scale hierarchical image database,''
\newblock in {\em CVPR}, 2009.

\bibitem{goyal2017accurate}
Priya Goyal, Piotr Doll{\'a}r, Ross Girshick, Pieter Noordhuis, Lukasz
  Wesolowski, Aapo Kyrola, Andrew Tulloch, Yangqing Jia, and Kaiming He,
\newblock ``Accurate, large minibatch sgd: Training imagenet in 1 hour,''
\newblock {\em arXiv:1706.02677}, 2017.

\bibitem{zhang2014part}
Ning Zhang, Jeff Donahue, Ross Girshick, and Trevor Darrell,
\newblock ``Part-based r-cnns for fine-grained category detection,''
\newblock in {\em ECCV}, 2014.

\bibitem{branson2014bird}
Steve Branson, Grant Van~Horn, Serge Belongie, and Pietro Perona,
\newblock ``Bird species categorization using pose normalized deep
  convolutional nets,''
\newblock {\em arXiv preprint arXiv:1406.2952}, 2014.

\bibitem{lin2015bilinear}
Tsung-Yu Lin, Aruni RoyChowdhury, and Subhransu Maji,
\newblock ``Bilinear cnn models for fine-grained visual recognition,''
\newblock in {\em ICCV}, 2015, pp. 1449--1457.

\bibitem{li2017dynamic}
Zhichao Li, Yi~Yang, Xiao Liu, Feng Zhou, Shilei Wen, and Wei Xu,
\newblock ``Dynamic computational time for visual attention,''
\newblock in {\em ICCV Workshops}, 2017.

\bibitem{lin2015deep}
Di~Lin, Xiaoyong Shen, Cewu Lu, and Jiaya Jia,
\newblock ``Deep lac: Deep localization, alignment and classification for
  fine-grained recognition,''
\newblock in {\em CVPR}, 2015.

\bibitem{krause2015fine}
Jonathan Krause, Hailin Jin, Jianchao Yang, and Li~Fei-Fei,
\newblock ``Fine-grained recognition without part annotations,''
\newblock in {\em CVPR}, 2015.

\bibitem{zhang2016spda}
Han Zhang, Tao Xu, Mohamed Elhoseiny, Xiaolei Huang, Shaoting Zhang, Ahmed
  Elgammal, and Dimitris Metaxas,
\newblock ``Spda-cnn: Unifying semantic part detection and abstraction for
  fine-grained recognition,''
\newblock in {\em CVPR}, 2016.

\bibitem{liu2016fully}
Xiao Liu, Tian Xia, Jiang Wang, Yi~Yang, Feng Zhou, and Yuanqing Lin,
\newblock ``Fully convolutional attention networks for fine-grained
  recognition,''
\newblock {\em arXiv:1603.06765}, 2016.

\bibitem{zheng2017learning}
Heliang Zheng, Jianlong Fu, Tao Mei, and Jiebo Luo,
\newblock ``Learning multi-attention convolutional neural network for
  fine-grained image recognition,''
\newblock in {\em ICCV}, 2017.

\end{thebibliography}

\end{document}